\documentclass[11pt]{article}

\usepackage[final]{acl}

\usepackage{times}
\usepackage{latexsym}

\usepackage[T1]{fontenc}
\usepackage[utf8]{inputenc}

\usepackage{microtype}

\usepackage{inconsolata}

\usepackage{graphicx}
\usepackage{amssymb}
\usepackage{amsmath}
\usepackage[most]{tcolorbox}

\newcommand{\lucie}[1]{{\color{black}#1}}
\usepackage[normalem]{ulem}

\usepackage{nicefrac}
\usepackage{tikz}
\title{On the Role of Unobserved Sequences on \\ Sample-based Uncertainty Quantification for LLMs}

\author{Lucie Kunitomo-Jacquin \\
  National Institute of Advanced Industrial\\ Science and Technology (AIST), 
Tokyo, Japan \\
  \texttt{kunitomo-jacquin.lucie@aist.go.jp} \\\And
  Edison Marrese-Taylor \\
  National Institute of Advanced Industrial\\ Science and Technology (AIST), Tokyo, Japan \\
  Graduate School of Engineering, \\The University of Tokyo, Tokyo, Japan \\
   \texttt{edison.marrese@aist.go.jp} \texttt{emarrese@weblab.t.u-tokyo.ac.jp} \\ \\
    \AND
  Ken Fukuda \\
  National Institute of Advanced Industrial Science and Technology (AIST) \\
   \texttt{ken.fukuda@aist.go.jp} \\
   }

\author{Lucie Kunitomo-Jacquin$^{1}$ \and Edison Marrese-Taylor$^{1,2}$ \and Ken Fukuda$^{1}$ \\
       $^{1}$ National Institute of Advanced Industrial Science and Technology (AIST),  Tokyo, Japan\\
       $^{2}$ Graduate School of Engineering, The University of Tokyo, Tokyo, Japan\\
        \texttt{kunitomo-jacquin.lucie@aist.go.jp} $\;$  \texttt{edison.marrese@aist.go.jp} \\ \texttt{emarrese@weblab.t.u-tokyo.ac.jp} $\;$  \texttt{ken.fukuda@aist.go.jp}
        }

\begin{document}
\maketitle
\begin{abstract}
Quantifying uncertainty in large language models (LLMs) is important for safety-critical applications because it helps spot incorrect answers, known as hallucinations. One major trend of uncertainty quantification methods is based on estimating the entropy of the distribution of the LLM's potential output sequences. This estimation is based on a set of output sequences and associated probabilities obtained by querying the LLM several times. In this paper, we advocate and experimentally show that the probability of unobserved sequences plays a crucial role, and we recommend future research to integrate it to enhance such LLM uncertainty quantification methods.
\end{abstract}

\section{Introduction}
The advent of large language models (LLMs) has revolutionized numerous fields by demonstrating remarkable capabilities across a diverse array of tasks. However, despite their impressive performance, these models often struggle with reliability issues, particularly due to factual inaccuracies in their outputs.  In this context, quantifying their confidence and adjusting them for various tasks can reduce risks and enhance the quality of outputs. 

However, uncertainty quantification (UQ) on LLMs remains challenging since the output possibilities for these models are substantially greater than those of discriminative models. As the generation length increases, the number of potential outcomes grows exponentially, making it unfeasible to evaluate all possible answers \cite{geng-etal-2024-survey}.  We can distinguish two types of uncertainty in LLMs: aleatoric uncertainty, stemming from inherent randomness, and epistemic uncertainty, resulting from a lack of information \cite{kendallWhatUncertaintiesWe2017}. Following previous work, we aim to quantify a measure of total uncertainty, i.e., aleatory and/or epistemic, as both types of uncertainty contribute to model errors.

Among the methods of uncertainty quantification for LLMs, we identify black-box methods, which assume access only to the generations, and white-box methods, which also utilize internal states of the LLM or token-level probabilities. In this paper, we focus on the latter, utilizing token-level probabilities. Concretely, we study sampling-based estimation methods, that is, approaches that rely on information (e.g. probability) obtained from multiple answers generated by the LLM, in order to quantify uncertainty.

Sample-based uncertainty  \lucie{quantification} methods via entropy \lucie{estimation}, like Predictive Entropy (\textsc{E}) \cite{malinin2020uncertainty} and the recently proposed Semantic entropy (\textsc{SE})  \cite{kuhn2023semantic,farquhar2024detecting}, have succeeded recently perhaps due to their simplicity, as they do not require any special training or architectural modifications. However, we note that these methods are themselves subject to epistemic uncertainty, as they rely on only a glimpse of the probability distribution of possible answers due to practical constraints. We highlight that methods like \textsc{E} and \textsc{SE}, in particular, do not account for this epistemic uncertainty, as they only consider the estimated probability of sampled sequences, neglecting the remaining probability of possible but unobserved answers.  

Recent work by \citet{abbasi2024believe} has moved in a similar direction and explored the concept of missing mass in UQ. However, their approach directly compares the distributions of the generated answers against the ground truth. Instead, here we present work focusing on modeling the probability of unobserved answers without the need for ground truth. Concretely, our aim is to propose a framework that enables us to incorporate this probability into existing formulations for estimation based on entropy. We provide technical considerations for the calculation of such probability and evaluate the relevance of one such implementation by using it as a UQ method.

\section{Proposed Approach}
\label{sec:background}
Let us denote by $x$ the object about which we quantify uncertainty; in our case study, $x$ refers to the input given to the LLM which often consists on a question and potentially a prompt.
\lucie{We denote by $(\mathcal{S}, \mathbb{P})$ the probability space, where $\mathcal{S}$ is the set of all possible sequences, and $\mathbb{P}$ is the probability measure over $\mathcal{S}$. } 
The entropy for the random output sequence of the LLM and the input $x$ is defined as Equation \ref{eq:Entropy} shows, below, where $p(s|x)$ is the probability of the sequence $s$ conditioned on the input $x$.
\begin{equation}
    E^*(x)=-\sum_{s \in \mathcal{S}} p(s|x) \log p(s|x).\label{eq:Entropy}
\end{equation}
As it is not realistic to compute the probability of all answers in $\mathcal{S}$, entropy-based UQ methods for LLMs estimate $E^*(x)$ on a set of $M$ sequences sampled from the model \lucie{denoted $s_1, \dots, s_M$}. Let us denote $A \subset \mathcal{S}$ the set of \lucie{unique} sampled answers and note that $|A| \leq M$ because some identical answers might be sampled multiple times.
Each answer $s \in A $ consists of a sequence of length $N$ in the set of vocabulary tokens $\mathcal{T}$. The probability of $s = (t_1, \dots t_N)$ is obtained by the product of conditional token probabilities via the language model, as follows.
\begin{equation}
    p(s|x) = \prod_i p(t_i|t_{<i}, x). \label{eq:proba_s}
\end{equation}
Some works have considered adjusting the calculation of sequence probabilities to account for varying sequence lengths. This is due to the tendency for longer sequences to exhibit lower joint likelihoods. To address this, a length normalized probability, which we denote $p'$ was proposed \cite{malinin2020uncertainty} as follows.
\begin{equation}
   \log p'(s|x) = \dfrac{1}{N} \sum_i  \log  p(t_i|t_{<i}, x). \label{eq:ln}
\end{equation}

We now focus on the probability of sequences not observed in the set $A$ of sequences provided by the LLM for a given input $x$. This probability is given by
\begin{eqnarray}
\mathbb{P}(\bar{A}| x)  &=& 1- \mathbb{P}(A | x) \\
                        &=& 1- \sum_{s \in A} p(s|x), 
\end{eqnarray}
where $\bar{A}$ denotes the complement set of $A$. \\

We believe that the probability of unobserved sequences can capture some of the uncertainty associated with an input $x$. When uncertainty is low, the model's output probabilities tend to be higher, leading to a lower probability for the unobserved sequences. Conversely, when uncertainty is high, the model's output probabilities are lower, resulting in a higher probability for the unobserved sequences.
In case of maximum uncertainty, all sequences in $\mathcal{S}$ are equally likely, with each having a probability of $\nicefrac{1}{|\mathcal{S}|}$. As a result, $\mathbb{P}(\bar{A}| x) = 1 - \nicefrac{M}{|\mathcal{S}|}$ approaches 1, especially when the \lucie{set of possible sequences} is very large. Conversely, in situations of minimal uncertainty, $\mathbb{P}(\bar{A}| x) = 0$.

In practice, we have two technical concerns related to the accurate calculation of probabilities for unobserved answers. Firstly, to the best of our knowledge, it is not always clear whether the last token, specifically the end-of-sequence (EOS) token, is considered in sequence probability calculations presented in Equation \ref{eq:proba_s}. If sequences do not include the EOS token, this raises concerns about the construction of the sample space, as two unfinished sequences are not mutually exclusive. Let us introduce a small example to illustrate our discussion about the sequence probability calculation. 

\begin{figure}
\begin{tikzpicture}[
  grow=right, 
  level distance=2cm, 
  sibling distance=2cm, 
  every node/.style={align=center},  
  edge from parent/.style={draw, ->, thick}
]
\scriptsize
  \node[align=center] {}
    child {node {vatican} 
      child {node {city} 
        child {node {EOS} edge from parent node[midway, above] {1}} edge from parent node[midway, above] {0.4}}
      child {node {EOS} edge from parent node[midway, above] {0.6}} edge from parent node[midway, above] {0.8}}
    child {node {$\bar{A}$} edge from parent node[midway, above] {0.2}};
\end{tikzpicture}
    \caption{Example of tree of possible sequences with token conditional probabilities.}
    \label{fig:tree}
\end{figure}
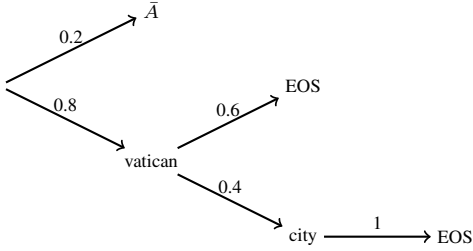

\paragraph{Example.} For the question input $x=$``\textit{Where are St. Peter's Basilica and the Sistine Chapel?}'', let us assume we observed two output sequences such that $A = \{ \text{``\textit{vatican}''}, \text{``\textit{vatican city}''} \}$ and consider the token conditional probabilities presented in Figure \ref{fig:tree}.  If we do not include the end-of-sequence token, the probability value of $0.8$ may be incorrectly interpreted as the probability of the sequence ``\textit{vatican}''. In fact, this represents the probability that the sequence starts with ``\textit{vatican}'', which also includes the possibility of the sequence being ``\textit{vatican city}''. Essentially, the events of the sequence beginning with ``\textit{vatican}'' and ``\textit{vatican city}'' are not mutually exclusive.

In addition to this issue, we also note that sequence length normalization techniques as shown in Equation \ref{eq:ln}, and often used approaches like \textbf{SE}, can distort probabilities, potentially leading to the sum of output probabilities differing from 1.

Due to the issues discussed above, we highlight that we cannot properly estimate the probability of unobserved answers with the usually-adopted sequence probability calculations. Thus, we compute the probability of sequences without sequence length normalization and considering the EOS token. Formally, we consider the probability of a sequence $s = (t_1, \ldots, t_N, \text{EOS})$ as 
\begin{equation}
p(s|x) = \prod_{i} p(t_i|t_{<i}, x) \times p(\text{EOS}|t_{\leq N}, x).\label{eq:pro_seq_with_eos}
\end{equation}  

\paragraph{Example (revisited).} Looking back at our previous example, if we consider the EOS token in the computation of the probability, we obtain $p(\text{``\textit{vatican}''}|x) = 0.8 \times 0.6= 0.48$, $p(\text{``\textit{vatican city}''}|x) = 0.8 \times 0.4 \times 1 = 0.32$ and the probability of the unobserved samples is $\mathbb{P}(\bar{A}|x) = 1-0.48-0.32 = 0.2$.

Based on this framework, here we present an alternative method for computing the uncertainty of an LLM where we directly use the value $\mathbb{P}(\bar{A}| x)$. We note that this approach, which we call \textit{Unobserved Probability} (\textsc{UP}), is arguably a very simple way to capture some part of the LLM uncertainty, as derived from our analysis.
 
\begin{itemize}
\item EOS-Inclusive \textsc{UP} (\textsc{EOS-UP}): this approach consist of quantifying the LLM uncertainty using $\mathbb{P}(\bar{A}| x)$ in the way we consider most suitable or recommended, i.e., accounting for the EOS token in calculating the sequence probabilities as in Equation \ref{eq:pro_seq_with_eos}.
\item Length-Normalized \textsc{UP} (\textsc{LN-UP}): we propose to quantify the LLM uncertainty using $\mathbb{P}(\bar{A}| x)$ as above, but considering the usual way for calculating the sequence probabilities, i.e., without accounting for EOS token and performing sequence length normalization, following Equation \ref{eq:ln}. 
\end{itemize}

\section{Experiments and Results}
\label{sec:exp}
In this section, we detail our experimental setup to evaluate the relevance of using the probability of unobserved answers for LLM uncertainty quantification via our proposed approach \textsc{UP}. We compare its performance with three entropy-based methods and also include, for reference, the probability of unobserved answers calculated using the conventional method for sequence probabilities.

\paragraph{Model and dataset.}
Our experiments focused on the uncertainty quantification for the \textit{falcon-40b-instruct} model \cite{almazrouei2023falcon} and were performed on a general knowledge dataset, TriviaQA \cite{joshi-etal-2017-triviaqa}. This model and dataset were recently used by \citet{nikitin2024kernel}. TriviaQA was also originally used by \citet{kuhn2023semantic} for their seminal work on \textsc{SE}.

\paragraph{Sampling.} 
We conducted our sampling using two styles of prompts. On the one hand, we adopt a prompt that pushes the model to produce short answers (\textsc{short}), ``\textit{Answer the following question as briefly as possible}''. This prompt was used on a more recent implementation of \textsc{SE}, presented by \citet{farquhar2024detecting}.\footnote{\url{https://github.com/jlko/semantic_uncertainty}} On the other hand, we also experiment with the original prompt (\textsc{normal}) presented by \citet{kuhn2023semantic}, and was also considered by \citet{nikitin2024kernel}, ``\textit{Answer the following question in a single brief but complete sentence.}''. Following the methodology of previous studies \cite{farquhar2024detecting,nikitin2024kernel}, we employed top-K sampling with \( K = 50 \) and nucleus sampling with \( p = 0.9 \) at a temperature of \( T = 1 \).

\paragraph{Evaluation Metric.}
In line with previous works \cite{farquhar2024detecting}, we evaluated the model's accuracy by sampling an additional answer at a lower temperature (\( T = 0.1 \)). Then we used another LLM, \textit{Meta-Llama-3-8B-Instruct} \cite{llama3modelcard}, to compare this answer with the ground truth answers from the datasets. \lucie{The prompts for checking answers correctness are provided in the appendix.} We evaluate uncertainty quantification methods by measuring their ability in predicting model output accuracy using the Area under the Receiver Operating Curve (AUROC). 

\paragraph{UQ methods.} We considered the following baseline methods in our experiments.
\begin{itemize}
\item Predictive Entropy (\textsc{E}) \cite{malinin2020uncertainty,kuhn2023semantic} is a \lucie{Monte-Carlo estimation} of predictive entropy, shown by Equation \ref{eq:e}, below. As per the original implementation, this uses sentence length normalization as in Equation \ref{eq:ln}. 
\lucie{
{\fontsize{9}{12}
\begin{equation}
    E(x) \approx - \dfrac{1}{M} \sum_{m=1}^M \log p'(s_m|x)
    \label{eq:e}
\end{equation}
}}
\vspace{-0.5cm}
% see eq (7) of Aichberger, L., Schweighofer, K., Hochreiter, S., 2024. Rethinking Uncertainty Estimation in Natural Language Generation. https://doi.org/10.48550/arXiv.2412.15176
\item Semantic Entropy (\textsc{SE}) \cite{kuhn2023semantic,farquhar2024detecting} is defined on a set of clusters capturing the distinct meaning, denoted $\mathcal{C}$. This consists in a sub-$\sigma$-algebra of the event-space of all possible answers $\mathcal{S}$. The uncertainty quantification is calculated by an approximation of the semantic entropy involving the normalization of the cluster probabilities \cite{farquhar2024detecting}, as shown in Equation \ref{eq:pprime} and Equation \ref{eq:se}, below, where $C\in \mathcal{C}$.
{\fontsize{9}{12}
\begin{eqnarray}
    p'(C|x) &=& \sum_{s \in C} p' (s|x) \label{eq:pprime}\\
    p''(C|x) &=& \dfrac{p'(C|x)}{\sum_{C\in \mathcal{C} } p'(C|x) }\\
    SE(x) &\approx&  - \sum_{c \in \mathcal{C}} p''(C|x) \log p''(C|x)  \quad \quad \label{eq:se}
\end{eqnarray}
}
\item Discrete Semantic Entropy (\textsc{DSE}) \cite{kuhn2023semantic,farquhar2024detecting} consists in a variant of \textsc{SE} where cluster probabilities are approximated by $p(C|x) \approx  \nicefrac{ |\{s \; : \; s\in C \} | }{M}$. 
\end{itemize}

\begin{figure}[t]
  \centering
  \includegraphics[width=\linewidth]{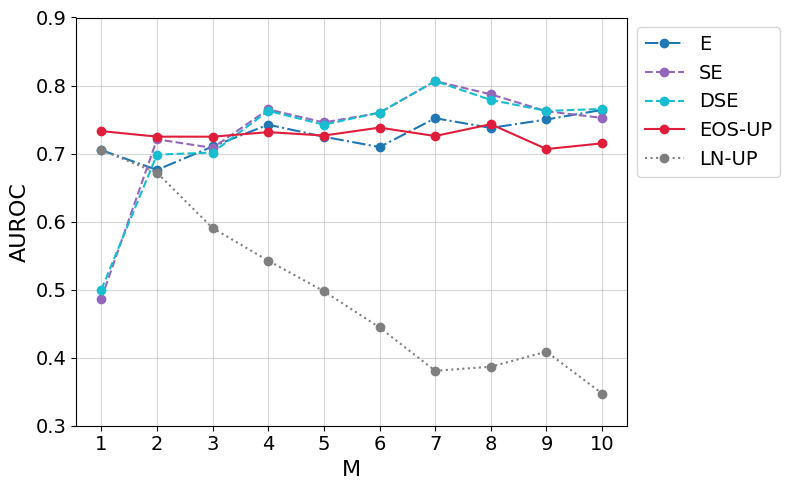}
  \includegraphics[width=\linewidth]{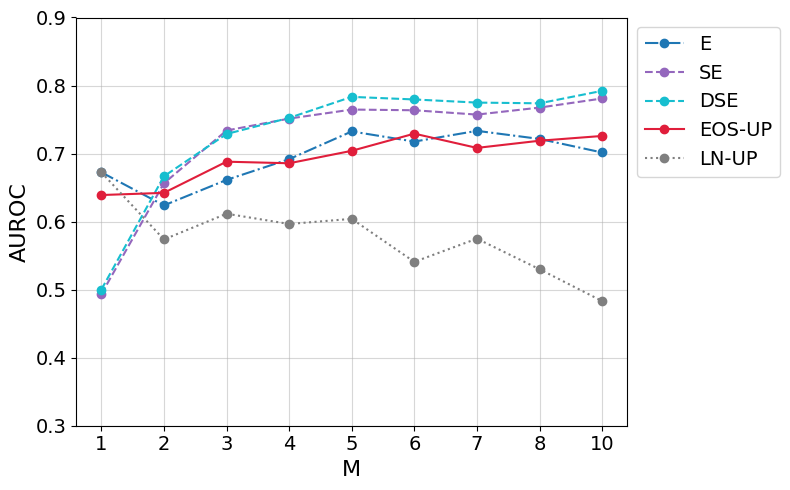}%\hfill
  \caption {Influence of the number of samples ($M$) for the LLM uncertainty quantification in terms of AUROC, for the \textsc{short} (top) and \textsc{normal} (bottom) answer length scenarios. We compare the performance of our proposed approach variations (\textsc{UP}) against relevant baselines. Results were computed on $500$ pairs of questions and ground truth answers on the \textit{falcon-40b-instruct} model.}
\label{fig:results}
\end{figure}

The results in terms of AUROC are presented in Figure \ref{fig:results}. We observe that the probability of unobserved answers \textsc{EOS-UP} is indeed relevant for quantifying uncertainty, achieving performance comparable to the Predictive Entropy (\textsc{E}) method. 

Moreover, we note that while state-of-the-art baselines (\textsc{E}, \textsc{SE}, and \textsc{DSE}) are affected by the number of available samples, the probability of unobserved answers maintains its performance even with a single sample. Sampling more answers from the LLM can generally lead to larger answer variability, and hence as $M$ grows, the effect of the probability of unobserved answers on the estimation decreases. Therefore, our results suggest that incorporating the probability of unobserved samples in the estimation of uncertainty can be of critical importance when the number of samples is limited (e.g. $M=1$). \lucie{Note that when $M=1$, $A = \{s_1\}$, \textsc{E} method reduces to $-\log p'(s_1|x)$, and \textsc{LN-UP} method to $1-p'(s_1|x)$. Since these quantity are strictly decreasing and monotonic with respect to $p'(s_1|x)$, they yield the same ranking over input instances and thus the same AUROC performance, as shown in Figure \ref{fig:results}.}

Finally, we observe the poor performance of our proposed probability of unobserved answers, considering length-normalization and no EOS token probability (\textsc{LN-UP}), not only remains the worst performing method for all $M$ values, but also that its performance decreases dramatically as $M$ grows. We think that, as shown by our technical considerations, our suggested way to compute this probability  (\textsc{EOS-UP}) is necessary to obtain an adequate estimation.

\section{Conclusion}
In this work, we aimed to focus on the probability of unobserved answers, which we note have been overlooked by existing entropy-based LLM UQ methods. We acknowledge that this probability captures only a portion of the uncertainty. For instance, hesitation between observed answers is not considered \lucie{since the probability of each separate observed answer is not used}. 

Our empirical results are encouraging and in the future we plan to integrate this quantity into existing entropy estimation methods. To achieve this, we believe a theoretical framework that considers both aleatoric and epistemic uncertainty, such as the Evidence Theory \cite{shafer1976mathematical,smets1994transferable} would be suitable.

We also note that current approaches of entropy-based UQ, %like \textsc{SE},
present other issues and limitations. Although the work of \cite{nikitin2024kernel} has made progress in this regard, we think further improvements are necessary, for example, by more directly modeling  hypernymy and hyponymy relationships across answers, and/or clusters of answers.

\section*{Acknowledgement} This paper is based on results obtained from a project, JPNP25006, commissioned by the New Energy and Industrial Technology Development Organization (NEDO).
\bibliography{bibliography}

\appendix
\section*{Appendix}  
\label{section:appendix}
To check the correctness of the answers, we used the same prompts as previous studies presented in Figure \ref{figure:axioms}.

\begin{figure}[h]
\begin{tcolorbox}[colback=gray!5!white,colframe=black!75!black,title=Prompt (single answer)]
\textit{We are assessing the quality of answers to the following question: \{\textbf{question}\} \textbackslash n
The expected answer is: \{\textbf{correct\_answer}\}. \textbackslash n The proposed answer is: \{\textbf{predicted\_answer}\} \textbackslash n Within the context of the question, does the proposed answer mean the same as the expected answer? \textbackslash n Respond only with yes or no.\textbackslash n Response:}\\
\end{tcolorbox}
\begin{tcolorbox}[colback=gray!5!white,colframe=black!75!black,title=Prompt (multiple answers)]
\textit{We are assessing the quality of answers to the following question: \{\textbf{question}\} \textbackslash n
The following are expected answers to this question: \{\textbf{correct\_answers}\}. \textbackslash n The proposed answer is: \{\textbf{predicted\_answer}\} \textbackslash n Within the context of the question, does the proposed answer mean the same as any of the expected answers? \textbackslash n Respond only with yes or no.\textbackslash n Response:}
\end{tcolorbox}
\caption{Prompts fed to the model in our experiments when providing a single (top) and many correct answers (bottom), where \textbf{placeholders} are denoted in bold.}
\label{figure:axioms}
\end{figure}

\end{document}